\documentclass[11pt]{article}

\usepackage[]{emnlp2021}

\usepackage{times}
\usepackage{latexsym}

\usepackage[T1]{fontenc}

\usepackage[utf8]{inputenc}

\usepackage{microtype}

\usepackage[normalem]{ulem}
\usepackage{booktabs}
\usepackage{multirow}
\usepackage{array}
\usepackage{csquotes}
\usepackage{amsmath}
\DeclareMathOperator*{\argmax}{arg\,max}

\newcommand{\legal}[1]{\textit{#1}}

\title{Automated Extraction of Sentencing Decisions from Court Cases in the Hebrew Language}

\author{
 Mohr Wenger$^{\diamondsuit}$\thanks{$\;\;$Equal contribution.} $\;\;$
 Tom Kalir$^{\diamondsuit*}$ $\;\;$
 Noga Berger$^{\spadesuit}$
 \\
 \bf
$^{\spadesuit\dagger}$Carmit Klar Chalamish $\;\;$
  Renana Keydar$^{\clubsuit\heartsuit} \;\;$
  Gabriel Stanovsky$^{\diamondsuit}$  \\
 $^{\diamondsuit}$School of Computer Science and Engineering
 $^{\clubsuit}$Faculty of Law 
 $^{\heartsuit}$Digital Humanities 
   \\
  The Hebrew University of Jerusalem,  Jerusalem, Israel \\
  $^{\spadesuit}$ The Association of Rape Crisis Centers in Israel \\
  $^{\dagger}$ Department of Criminology,
 Conflict Resolution, Management \& Negotiation Graduate Program, \\
Bar-Ilan University, Israel  \\
 \{mohr.wenger, renana.keydar, gabriel.stanovsky\}@mail.huji.ac.il
}
\usepackage{graphicx}
\usepackage{graphicx}
\usepackage{subcaption}
\usepackage[export]{adjustbox}
\usepackage{wrapfig}
\graphicspath{ {./images/} }
\begin{document}
\maketitle
\begin{abstract}
We present the task of Automated Punishment Extraction (APE) in sentencing decisions from criminal court cases in Hebrew. Addressing APE will enable the identification of sentencing patterns and constitute an important stepping stone for many follow up legal NLP applications in Hebrew, including the prediction of sentencing decisions. We curate a dataset of sexual assault sentencing decisions and a manually-annotated evaluation dataset, and implement rule-based and supervised models. We find that while supervised models can identify the sentence containing the punishment with good accuracy, rule-based approaches outperform them on the full APE task. We conclude by presenting a first analysis of sentencing patterns in our dataset and analyze common models' errors, indicating avenues for future work, such as distinguishing between probation and actual imprisonment punishment. 
We will make all our resources available upon request, including data, annotation, and first benchmark models.
\end{abstract}

\section{Introduction}
The legal world is rife with data, from constitutions and national legislation to legal cases and court decisions. Much of the legal data, however, comes in unstructured formats that pose critical challenges for extracting and analyzing it in systematic ways.
In addition, different countries vary in their legal systems, norms and conventions, further compounding the challenges in developing multilingual approaches~\citep{peruginelli2009accessing}.

While legal NLP is gaining traction in recent years~\cite{NLPtomodellegislation,dale2019law,how_does_NLP_help_legal}, relatively little attention has been given to low-resource settings outside of the English language, where the availability of tools such as large pretrained language models, syntactic parsers, or named entity recognizers is limited.

In this work, conducted as part of an on-going  collaboration with The Association of Rape Crisis Centers in Israel (\textbf{ARCCI}), we focus specifically on the task of Automated Punishment Extraction (\textbf{APE}) in sexual assault cases in Hebrew within Israeli court sentencing decisions (see formal task definition in Section~\ref{sec:task-def}).
Punishment decisions are of special importance as they constitute a prerequisite for many other downstream tasks in legal NLP and digital humanities, such as legal prediction of judicial decisions~\cite{aletras2016predicting,legal_predicitons2021} and detecting biases in court decisions~\cite{Bias_detection2020}.
APE is difficult in the Israeli court system. 
This is due to the fact that sentencing decisions for criminal offences are reported, in natural language idiomatic to the legal field, in the written sentencing decision. 
We focus on sexual assault cases due to the legal and public debate around claims of lenient punishments~\cite{phillips2020six}, that in the absence of systematic rigorous data collection cannot be empirically examined and assessed. This worldwide debate requires legal NLP methods in multiple languages and legal systems.\footnote{https://www.haaretz.com/israel-news/.premium-women-decry-lenient-rape-sentence-1.5383195, https://balkaninsight.com/2021/04/05/victims-discouraged-by-lenient-sentences-for-sex-crimes-in-serbia/.}


To address this challenge, we begin by curating a dataset of sexual assault sentencing decisions from the years 1990-2021 and manually annotate punishment in a subset of 100 cases with the use 
of legal experts in our team and in collaboration with ARCCI (Section~\ref{sec:data}). Following, in Section~\ref{sec:model}, we use this data to build several models for the APE task, including rule-based and supervised methods, based on linguistically and semantically informed features, setting first benchmark results on the APE task in Hebrew. We thoroughly analyze our models' performance in Section~\ref{sec:evaluation}, finding that they are capable of extracting the correct punishment in 68\% of the cases,
while the best model's average error is roughly 5 months, attesting to the difficulty of the task.
Based on our models, we find that in our data the median predicted punishment is 3 years, while more than a third of the punishments are below 15 months. Although these figures are obtained on a medium-size corpus, using automatic measures which do not account for the type of offense, we note that they are well below the maximum punishments for sexual offenses as determined by the Israeli legislator, which range between 2-7 years for indecent acts and sodomy and up to 20 years for aggravated rape.

We conclude by analyzing common error patterns in our models. For example, we find that models often tend to erroneously extract a probation imprisonment punishment instead of the actual imprisonment punishment. Distinguishing between the two is left as an interesting avenue for future work.
To the best of our knowledge, this is the first examination of automatic punishment extraction in the Hebrew language. It includes data collection, annotation, and benchmark models. We hope it will spur further research into this important task.

\section{Task Definition}
\label{sec:task-def}
We define the task of \textbf{APE} as the process of automatically extracting the punishment from the sentencing decision. In the Israeli legal system, the punishment is given in a separate sentencing decision, following a plea bargain or a guilty verdict. In the sentencing decision, the court can impose different types of punishment: imprisonment, probation, or community service. In addition, the court can also impose fines and order the defendant to pay restitution to the victim.
We consider all of the punitive elements mentioned as part of the APE process.
However, in this work, we focus on the extraction of the \textit{actual} imprisonment (i.e. jail time). Given the text of a sentencing decision, we first need to distinguish between the different types of punishment (imprisonment, probation, community work, fines, etc.); then, we need to extract only the sentence that relates the duration of the  \textit{actual} imprisonment penalty, i.e. the number of months or years in prison imposed on the defendant. This is particularly challenging since the court decision often includes both the duration of the actual imprisonment, as well as the duration of the conditional imprisonment (i.e. probation). Both are referred to in Hebrew using the same term ``Ma'asar'' (lit. imprisonment), and indicated by the same units of months and years. For example (translation and emphasis by the authors): 

\begin{quote}
 
 ``We impose on the defendant the following punishment: 48 months \textit{imprisonment}, of which the defendant will serve 30 months actual \textit{imprisonment} and the rest, 18 months, will be conditional \textit{imprisonment}...(CrimC 1124/04)''.

 \end{quote}
 
 In this case, the APE task is to extract ``30 months'' as the the actual imprisonment punishment. 
This also exemplifies the typical linguistic difficulty of the task --- The noun ``imprisonment'' repeats three times, referring first to the total punishment imposed, then to the actual imprisonment, and then to the probation.

\section{A Corpus of Annotated Sentencing Decisions in Israeli Law}
\label{sec:data}
This section describes the construction of our corpus, which to the best of our knowledge is the first annotated legal corpus of sentencing decisions in sexual assault cases in Hebrew. In Section~\ref{sec:text} we discuss the cases comprising the corpus, consisting of 30 years of sentencing decisions in sexual offense cases, and in Section~\ref{sec:annotation} we present the manual annotation schema of the different types of punishment and the duration (in months and years) of the actual or conditional imprisonment which the courts imposed in these cases.

\subsection{Data Collection}
\label{sec:text}

We compiled a corpus containing sentencing decisions from Israel Magistrate and District Courts, from the years 1990 - 2021, as collected by Nevo legal database.\footnote{\url{https://www.nevo.co.il}. The data does not represent all the cases that were held in court but only those that were documented in the Nevo database.} All the cases in the corpus deal with sexual offenses under sections 345-351 of the Israel Penal Law, 5737-1977, including offenses of rape, sodomy, indecent acts and sex offenses within the family.

\begin{figure}[tb!]
\centering
\includegraphics[width=\columnwidth]{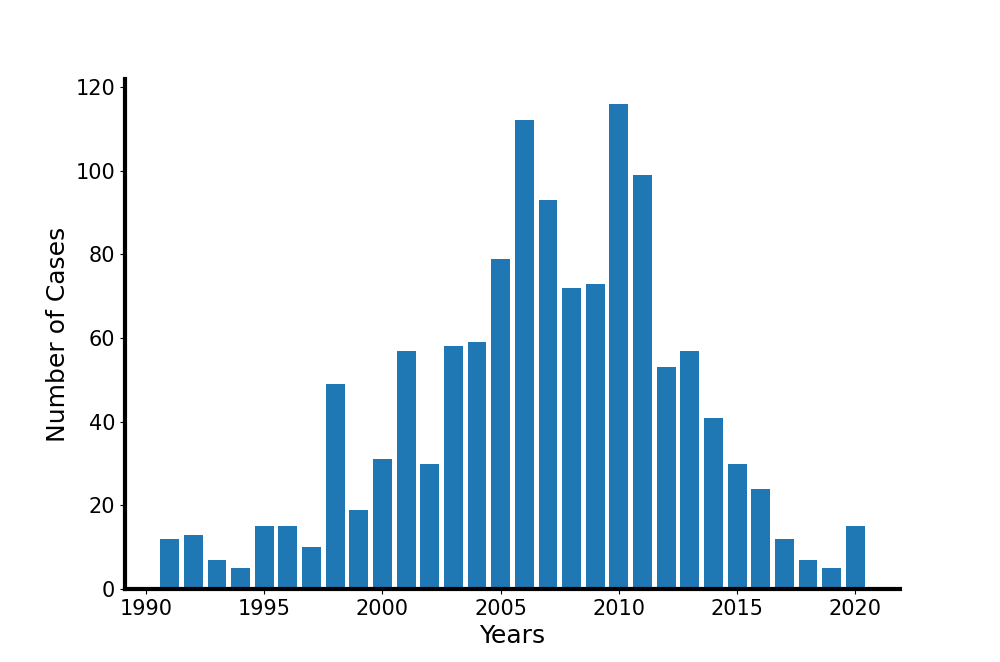}
\caption{Distribution of  legal cases in our corpus by year. The small number of cases before year 2000 is probably due to changes in digitization of legal documents.}
\label{fig:cases_years}
\end{figure}

The characteristics of our corpus are presented in Table~\ref{tab:data-stats} and Figure~\ref{fig:cases_years}.
This corpus, which is available upon request, directly lends itself for the quantitative exploration of sentencing and punishment patterns in sexual assault cases in the Israeli legal system, as well as for other areas of criminal law. In total this includes 1043 cases, 181k sentences and 3M words, of which we annotated a subset of 100 cases that include 13k sentences and 210k words. The sentences vary much in length with an average length of 16.5+-15.4 words in all files and 25+-16.5 words in the annotated subset.\footnote{The high variance in decisions' length in the legal domain is due in part to the difficulty in segmenting legal texts, as noted by~\citet{sanchez-2019-sentence}. }

\begin{table*}[]
\center
\begin{tabular}{l|l|l|l|c|c|c|}
\cline{2-7}
\multirow{2}{*}{}              & \multirow{2}{*}{\textbf{Cases}} & \multirow{2}{*}{\textbf{Sentences}} & \multirow{2}{*}{\textbf{Words}} & \multicolumn{3}{l|}{\textbf{Sentence Length (words)}} \\ \cline{5-7} 
                               &                                 &                                     &                                 & \textbf{Average length} & \textbf{Min} & \textbf{Max} \\ \hline
\multicolumn{1}{|c|}{All}      & \multicolumn{1}{c|}{1043}       & \multicolumn{1}{c|}{181K}          & \multicolumn{1}{c|}{3M}    & $16.5 [\pm 15.4]$              & 0            & 433          \\ \hline
\multicolumn{1}{|c|}{Annotated} & \multicolumn{1}{c|}{100}        & \multicolumn{1}{c|}{13K}          & \multicolumn{1}{c|}{210K}     & $25 [\pm 16.5]$               & 3            & 321          \\ \hline
\end{tabular}
\caption{Statistics of our annotated data, referring to the full corpus as well as to the annotated subset. In both cases, each legal decision contains many sentences and the length of the decisions varies considerably.\label{tab:data-stats}}
\end{table*}
\subsection{Annotation}
\label{sec:annotation}
We set out to annotate the punishment as defined in Section~\ref{sec:task-def} in a sample of 100 sentencing decisions. We achieve this in two semi-automatic annotation steps as exemplified in Table~\ref{tab:annots} and elaborated below. This setup was found to be useful both in terms of the annotation quality, as well as in providing direct supervision signal for the intermediate tasks. All annotations were done by legal scholars and practitioners or under their guidance and supervision.

\begin{table*}[]
\small
\centering
\begin{tabular}{p{5cm}p{8cm}p{0.8cm}}
\toprule
\textbf{English translation} & 
\textbf{Comments} & 
\textbf{Punish-ment} \\ \midrule

\legal{Therefore, the prosecution requested to impose a punishment of lengthy actual imprisonment, conditional imprisonment, and significant compensation to the victim.} 
 &
The prosecution's position on the desired punishment for the defendant. We attempt to rule it out based on the verb ``request'' and based on the fact that there are no numbers in this sentence.
 &
0  \\\midrule

\legal{Thus for example in CrimA 1049/12 Anon v The State of Israel the appellant was convicted of committing an indecent act on his 9-year-old granddaughter (...) and for that, he was sentenced to 12 months actual imprisonment, probation, and a fine of 40,000 NIS.} &
A reference in the decision to punishments that were imposed in prior cases, usually as example for standard of punishment. We attempt to  rule it out based on the past tense of the verb ``was sentenced''  and characters such as ``/''  that mark the docket number of a prior court case. &
0 \\\midrule

\legal{48 months imprisonment, of which the defendant will serve 30 months of actual imprisonment and the rest, 18 months, in conditional imprisonment.} &
Combined punishment statement consisting of an actual imprisonment and probation.  We attempt to extract only the number of months of actual imprisonment [30], while ruling out the total months [48] and the probation months [18]. 
In cases of combined punishment, we do not rule out the sentence. One way is to  check if it contains the term ``and the rest'', which indicates the actual imprisonment, preceding the term ``and the rest''  &
30 \\\midrule

\legal{Actual imprisonment will start
on 31.}& Procedural orders regarding the execution of the imprisonment. These are normally short sentences that include the word for imprisonment and a number (normally either date or hour), which render them very confusing for our models. In addition, in many cases they appear not as a full sentence but in fact truncated immediately after the first period due to limitations of sentence extraction. We attempt to rule them out by the fact that they do not contain a time unit.& 0
\\\midrule

\legal{A fine of 5,000 NIS or 30 days imprisonment instead.} & 
A fine that is given in addition to the actual imprisonment. The fine can be substituted by a 30-day imprisonment alternative. We attempt to rule it out based on the word ``fine'' that does not appear in a sentence reflecting the actual imprisonment. & 0
\\ \bottomrule

\end{tabular}
\caption{APE annotation example, including all the sentences in which the Hebrew word for imprisonment appears. We provide an example from our data for some of the challenges of this task. We refer to some of these examples once more in the models' error analyses in Section~\ref{sec:evaluation}. For brevity's sake we condense the two annotation phases into a single ``prison time'' column, which is marked zero if the sentence does not convey a punishment.}

\label{tab:annots}
\end{table*}

\paragraph{Imprisonment Sentence Identification.} This is a sentence level, binary annotation task, as exemplified in the third column in Table~\ref{tab:annots} (labelled ``Prison [Y/N]''). 
We identified that the actual imprisonment is often contained within a single sentence in the decision. Sometimes this sentence also contains the conditioned punishment, for example, see Table~\ref{tab:annots}, where  the third row shows the verdict of 48 months imprisonment, of which 30 months are actual imprisonment and the remaining 18 are conditioned. 
In other cases where the actual and conditional imprisonment are in separated sentences, we were interested in the actual imprisonment.
In Table~\ref{tab:annots} we see four different sentences that contain the word ``imprisonment'', however only the third sentence contains the actual imprisonment imposed on the defendant. 
In this case it contains also the conditioned imprisonment but this is not always the case, in section~\ref{sec:evaluation} we will see how this affects our models' performance.

Naturally, the vast majority of sentences should be labelled negatively, as most of the sentences do not convey the punishment.
To ease the annotation process, we automatically labelled as negative all sentences which did not contain a phrase from a predefined list of words indicating sentencing decisions and which were found to convey the punishment in our data. Each sentence was linked to its document, so that in cases of ambiguity we could evaluate the single sentence against the full judicial decision to reach a conclusive annotation. This resulted in negative annotation for 11.2K sentences in our dataset (85\%). The remaining sentences (15\%) were manually annotated  with the guidance and supervision of legal scholars. 

A major challenge in annotating these remaining sentences was differentiating between the punishment imposed on the defendant in this particular case and the discussion of previous punishments. For example, reference to punishments that were imposed on the defendant in previous cases or punishments that were given in similar cases which then serve to establish the punishment standard. These often use similar terminology to that of the current punishment, for example see the second row of Table~\ref{tab:annots}.
In other cases, a sentence containing the imprisonment in the current case is followed by a sentence which activates a previous probation. In such cases, we annotate both sentences as conveying an imprisonment.

This annotation step resulted in 132 sentences annotated positively with either actual imprisonment or probation, while the remaining 13K sentences were marked negatively, either automatically or by human experts. This annotation averaged 1.26 sentence marked positive for conveying the punishment per case, thus matching our intuition that the punishment in each decision tends to be conveyed in a single sentence.

\paragraph{Imprisonment Time Annotation.}
In the second stage, we manually annotate an integer denoting the duration (in months) of imprisonment, as exemplified in the last column in Table~\ref{tab:annots} (denoted ``Prison time'').
We presented our annotators with the sentences found in the previous stage to contain an imprisonment punishment, and ask them to label the number of actual imprisonment time in months. For example, the third sentence in Table~\ref{tab:annots} is annotated with 30 months of imprisonment. Overall, punishments vary between 0 months (no actual imprisonment) to 168 months (14 years actual imprisonment). The average punishment was 32 months, and the median was 15 months.

\section{Models}
\label{sec:model}
\begin{figure*}[]
\includegraphics[width = \textwidth]{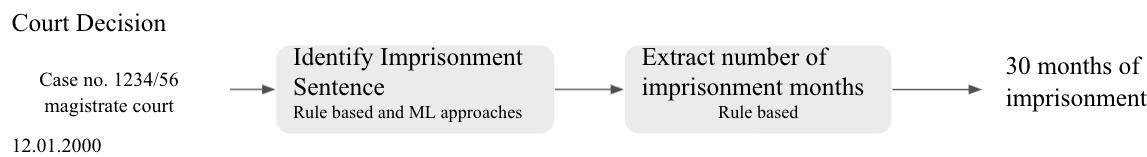}
\caption{
High-level diagram of our models for extracting duration of imprisonment from court decisions (left) to imposed punishment (right). We begin by identifying candidate sentences for containing the imprisonment  (Section~\ref{sec:filtering}), followed by an extraction of the imprisonment term, in months (Section~\ref{sec:months}).
\label{model_plot} }
\end{figure*}

We present several models for predicting the imprisonment incurred in free text sentencing decisions.
The high-level approach, depicted in Figure~\ref{model_plot}, is composed of two steps, following the human annotation process, described in the previous section. 
First, we identify sentences containing the imprisonment punishment~(Section \ref{sec:filtering}), from which we extract the 
term itself, and normalize to number of imprisonment months~(Section~\ref{sec:months}).

\subsection{Imprisonment Sentence Detection}
\label{sec:filtering}
First, we try to find the sentences conveying the imprisonment.
We start with a keyword based approach to filter a subset of relevant sentences (e.g., the Hebrew word for imprisonment). 
This allows us to reduce the number of sentences per case from hundreds or even thousands to approximately 14 sentences per case on average. 

Following, we aim to extract one sentence predicted to contain the imprisonment. We experiment with a rule-based approach, and several machine learning models, including SVM and random forest.
In all models we use linguistic as well as structural document-level features, such as the position of the sentence within the document.

\paragraph{Rule-based approach.}
This approach consists of a scoring system for several keywords, compiled based on the authors' legal expertise. Specifically, we created four lists: two lists with strong and moderate words that indicate this is the target sentence, hence ``positive words'', and two lists including strong and moderate words that indicate the sentence probably does \textbf{not} include the imprisonment decision in the case, hence ``negative words'', 
Each of these were heuristically scored based on a held-out development set. A sentence was deemed positive if and only if its score surpasses a threshold, determined as well based on the development set. See the full details in our codebase, to be made available upon publication.

\begin{itemize}
\item Strong positive words: verbs that indicate the judicial decision on the punishment, such as ``sentencing'', ``deciding'', ``imposing'' etc., all in present tense.
\item Moderate positive words: include the infinitive form of the strong positive words. In Hebrew, these can be used both as past and present tense, which is why we decided to score these moderately, in case these were used to refer to past decisions, which the judge uses to establish the punishment standard.
\item Moderate negative words: characters such as brackets and backslash that indicate a reference to a docket number, usually of a previous legal case.
\item Strong negative words: Hebrew words relating a request or petition brought before the court regarding the desired punishment, usually by one of the sides, as opposed to the final judicial decision which is an order of the court.
\end{itemize}

 \paragraph{Supervised modeling.}
This approach consists of using features similar to the rule-based approach, and experimenting with different machine learning techniques for determining their weights. This is divided to two stages: \textit{Stage 1} -  identifying punishment sentences by assigning probabilities and choosing all sentences above a threshold. 
\textit{Stage 2} - extracting a single sentence in each document, which includes the actual imprisonment, since our final goal is extracting the number of imprisonment months. For this we perform an argmax over the probabilities assigned by the model:

\begin{equation}
    M(D) = \argmax_{s \in D} P_{\theta_M}(s)
    \label{eq:framework}
\end{equation}

Where $D$ is a legal case, composed of a list of sentences $s$, $M$ denotes different models whose weights are denoted with $\theta_M$, and $M(D)$ is the predicted sentence from case $D$ according to model $M$.
Within this framework, we compare two models: support vector machine (\textbf{SVM}; \citealt{cortes1995support}) and random forest (\textbf{RF}; \citealt{ho1995random}), both trained and tested using cross-validation on the annotated subset.

\subsection{Extracting the number of months of imprisonment}
\label{sec:months}
After identifying a candidate sentence for imprisonment, we implement the following pipeline to extract the number of months of imprisonment incurred.

\paragraph{Identifying numbers in Hebrew.}
First, we identify all candidate numbers in the sentence. To achieve this we use a regular expression for each digit, as well as rules for converting multiple-digit numbers. 

The Hebrew number, similar to English, has a basic form that appears if it is between 0 - 10, an indicator for if it is between 10-20 and a slightly different form for if it is a multiplication of ten (i.e., twenty, thirty). However, there are also several differences that pose challenges unique to Hebrew:
\begin{itemize}

\item \textit{Suffixes that are range dependant -} we introduced different rules to account for Hebrew morphology in different number ranges.
\item \textit{One year/ One month punishment can be deduced only by elimination - } Hebrew does not have a non-specific determiner, so a punishment of one year for example can be phrased as ``year of imprisonment'', without a time unit (``one'') or a determiner. Similarly, for any number above twenty, it is accepted in Hebrew to mention the time unit in single form rather than plural form (``20 year''). 
We overcome this challenge by using elimination for determining the found time unit was an indicator for 1 year / 1 month.
\item \textit{Spelling variation - } 
Spelling in Hebrew often varies due to its treatment of vowels, which are sometimes indicated with diacritics (Niqqud), or omitted altogether~\cite{ravid2006vowel}. To address this, we account for all the possible vowels and syllable combination for all digits.

 \end{itemize}
 \paragraph{Identifying the imprisonment duration.}
 Following, we aim to find the number that indicates the length of actual imprisonment. First, we checked the following heuristic - is this sentence of the form ``The total sentence of Z units of time, which consists of X actual imprisonment time and Y conditioned imprisonment'' (see row 3 of table ~\ref{tab:annots}), if so, we return the value X. This was done by checking if there are exactly three numbers mentioned, and if Z=X+Y. If that is in fact the case, then we return X as the actual imprisonment. Otherwise,  we created a scoring method, which looks for certain features, such as the distance between the number and some time unit indicator such as ``years'' or ``months'', or the distance between the number and the end of the document. This was also locally assessed, i.e. there was no absolute threshold and the number with the best score within each sentence was chosen.

\section{Evaluation}
\label{sec:evaluation}

\begin{table}[tb!]
\small
\resizebox{\columnwidth}{!}{  
\begin{tabular}{lccc}
\toprule
    Model &Sent F1 & APE  F1 & Avg Err (Months) \\ \midrule
\textbf{Rule-based}   & \textbf{0.68}   & \textbf{0.65}   & \textbf{5}   \\ 
\textbf{RF}  & 0.61                                                                     & 0.58                                                                         & 11.6                                                                    \\ 
\textbf{SVM} & 0.54                                                                     & 0.5                                                                          & 10                                                                      \\ \bottomrule
\end{tabular}
}
\caption{Full task evaluation using the sentences extracted by the different models. It shows that in the overall task the rule based models achieved the highest accuracy and also the lowest average distance from the ground truth of actual imprisonment (manually tagged).} 
\label{tab:full_APE}
\end{table}

In this section we analyze the performance of the models described in section~\ref{sec:model} in the different stages of the task.
The main results are presented in Table~\ref{tab:full_APE}. In addition, we examine supervised model performance on sentence identification in Table~\ref{tab:model_comp}, perform manual error analysis in Table~\ref{tab:joint_err}, and plot punishment trends on the entire corpus in Figure~\ref{fig:cases_years}.
We draw several conclusions based on these results. 

\paragraph{The supervised models present high accuracy in the sentence identification task. }
 Both supervised models in Table~\ref{tab:model_comp} show high recall in tagging the sentences which convey the  imprisonment. However, they also tag some additional false positive sentences, hence decreasing the precision rate. In total, this results in between 2 and 5 sentences tagged as positive in each document.
 
 \begin{table}[tb!]
\small
\center
\begin{tabular}{lllll}
\toprule
Model                                                             & Recall &Precision & F1    \\ \midrule 
SVM                                                               & 0.9 & 0.86 & 0.88 \\  
RF                                                                & 1 & 0.8  & 0.89 \\ 
\bottomrule
\end{tabular}
\caption{A comparison between the first stage of the supervised approaches in their ability to identify the sentence that includes the punishment. Note this is the ability to extract 2 - 5 sentences of which only one is correct, for the full APE task we need to choose the correct one.}
\label{tab:model_comp}
\end{table}

\paragraph{The supervised models' probability is not well-calibrated.} The APE task requires choosing one sentence from which the number of months of actual imprisonment is extracted in the next stage. This means that for each false positive sentence there is exactly one equivalent false negative. Hence, in our case precision equals recall. In the rule-based approach we noticed that a different threshold applies for each legal case, hence we scored them separately. Our assumption was that the supervised models would score similar features to those used in the rule-based in a more accurate way. Thus we attempted a similar local threshold approach, by performing argmax on the learner probability of all sentences from the same legal case, as defined in Equation~\ref{eq:framework}. However this was not the case, as observed in Table~\ref{tab:full_APE}, the rule-based approach achieves better results in extracting a single sentence and also in extracting the actual imprisonment time. This points to the supervised models' probability not being calibrated, perhaps due to the low resrouce domain and small number of samples.

\begin{table*}[h!]
\small
\center
\begin{tabular}{lllll}
\toprule
\multirow{2}{*}{Error Type} & \multicolumn{3}{c}{\% of examined cases}                                                        & \multirow{2}{*}{Example (English Translation)}                                                                                   \\ 
                            & \multicolumn{1}{c}{\textbf{RB}} & \multicolumn{1}{c}{\textbf{SVM}} & \textbf{RF}               &                                                                                                                                  \\ \midrule
Probation                   & 36\%                             & 37.5\%                            & \multicolumn{1}{l}{\textbf{65\%}} & \textit{In addition, 18 months probation.}                                                                                       \\ 
Ref to prev. case & \textbf{46\%}                             & \multicolumn{1}{c}{-}            & \multicolumn{1}{l}{18\%} & \textit{Sentenced to 12 months actual imprisonment.} \\ 
Fines                       & \multicolumn{1}{c}{-}           & \textbf{27.5}\%                            & -                         & \textit{A fine of 5,000 NIS or 30 days imprisonment instead.}                                                                    \\ 
Procedural          & \multicolumn{1}{c}{-}           & \textbf{17.5\%}                            & -                         & \textit{Imprisonment starts at 31.}                                                                                               \\ \bottomrule
\end{tabular}
\caption {Error analysis for all three models (Rule-Based, Support-Vector Machine, and Random Forest). We used for imprisonment sentence detection. Sentences that contain probation are the only common cause for errors in all models. In both supervised approaches they are also responsible for the highest percent of errors. For the rule based we observe that references to past cases were more confusing, this was also confusing for the manual tagging task. In total the RF and rule based were more similar in their error analysis than the SVM. The remaining errors for each classifier did not fall under a common category and could be generally defined as miscellaneous.  }
\label{tab:joint_err}
\end{table*}

\paragraph{All models tend to confuse probation with actual imprisonment.}
Error analysis in Table~\ref{tab:joint_err} shows that the most common error was extracting a sentence with the probation rather than the actual imprisonment. We remind that ``probation'' in Hebrew is phrased ``conditional imprisonment'', which may lead to this confusion. In many cases, probation and actual imprisonment are pronounced in one sentence. In other cases, the probation directly follows the pronouncement of the actual imprisonment, and has similar syntactic and semantic cues.

\paragraph{Other error patterns.}
Rule-based and RF errors are similar, containing mostly references from past cases. This type of error includes sentencing decisions either of similar crimes or past cases of the defendant (see Table~\ref{tab:annots}, row 2). These sentences are similar in structure to those reflecting the actual imprisonment, and also confused legal expert annotators. In contrast, SVM errs on extracting sentences describing fines (accompanied by an alternative of imprisonment) or procedures regarding the execution of the incurred imprisonment rather than sentences reflecting actual imprisonment duration. While those cases include a number and the Hebrew word for imprisonment, they are easily ruled out by human annotators. Both learners use this word as a feature, however SVM still makes mistakes classifying fines.

\paragraph{Inter-annotator agreement reveals the limitations  of the sentence-level approach.} We asked legal experts to evaluate the rule-based performance by rating each sentence that the rule-based model predicted as whether it reflects the actual imprisonment or not. This is the same task the models were required to perform. We  used Cohen's kappa to measure the inter annotators agreement for each pair, presented at table~\ref{tab:cohen_k}  and Fleiss' kappa \cite{fleiss1971measuring} for measuring the level of agreement between five different annotators and three different classes (sentence is indicative of punishment/is not indicative of punishment/cannot decide). The annotators achieved a score of 0.341 which is considered a fair agreement~\cite{kappa2005understanding}. On average the taggers managed to correctly find the actual imprisonment in 79\% of the sentences. In many cases the annotators expressed doubt regarding their ability to tag the sentence as the actual imprisonment solely based on the single sentence extracted by the algorithm, without the context of the full judicial decision. This suggests a limit to the ability to determine whether a sentence from a legal document contains an actual imprisonment time, without the larger document context.

\begin{table}[]
\center
\begin{tabular}{l|ccccc}
     & ann1 & ann2 & ann3 & ann4 & ann5 \\ \midrule
ann1 & -  & 0.13 & 0.46 & 0.36 & 0.34 \\
ann2 & 0.13 & -  & 0.14 & 0.31 & 0.42 \\
ann3 & 0.46 & 0.14 & -  & 0.48 & 0.34 \\
ann4 & 0.36 & 0.31 & 0.48 & -  & 0.63 \\
ann5 & 0.34 & 0.42 & 0.34 & 0.63 & - 
\end{tabular}
\caption{Agreement between each pair of annotators in terms of Cohen's Kappa. The 5-way agreement between all annotators is 0.341 Fleiss' kappa.}
\label{tab:cohen_k}
\end{table}

\paragraph{There is room for improvement in extracting the actual imprisonment sentence.} 
Table~\ref{tab:full_APE} demonstrates the different models' performances in the full APE task. In this case, the rule-based approach performs best with an average error of 5 months, while supervised models reach an average error of 10-11.6 months. This table also shows how sentence extraction accuracy alone does not predict the ability to succeed in APE task. This is also affected by the type of mistakes, i.e. when the wrong sentence is predicted, the number of months extracted is not directly related to the actual imprisonment. 
However, given the correct sentence, extracting the duration of imprisonment was accurate in 89.7\% of the cases. Therefore improvements could be achieved by better extraction of the actual imprisonment sentence. Future work may consider separately tagging the probation sentence, as its structure might be easier for the model to learn. Once learned, it could be used as an anchor. This problem could also benefit from employing contextualized representations such as adapting a Hebrew language model, such as Alephbert~\cite{seker2021alephberta} to the legal domain, an approach recently shown effective in English~\cite{2020legalBERT}.

\paragraph{A post-hoc summary of the legal decision improves performance.} Nevo, the legal database we use, provides a ``mini ratio'', a post-hoc summary of each decision in a few sentences, written by Nevo's editorial team. When we add this mini ratio to the annotation, it increases the supervised models' ability to extract the target sentence by about 50\%, showing that shorter inputs lead to better generalization on small-scale datasets.

\begin{figure}[tb!]
    \centering
\includegraphics[width = \columnwidth]{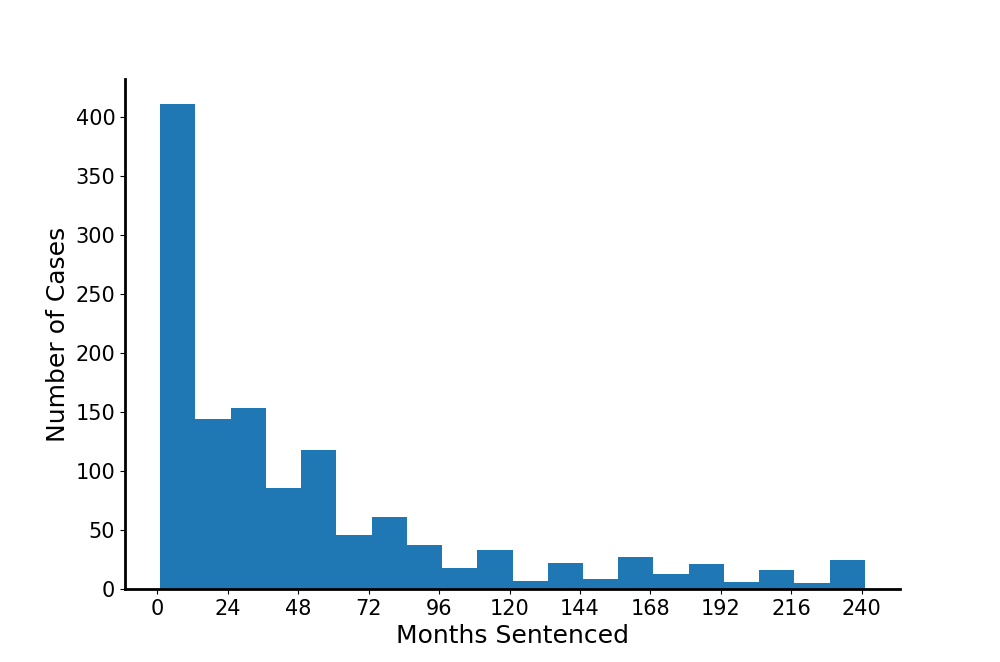}
\caption{The distribution of predicted punishments in our corpus. These were extracted using our rule-based model which performed best on this task. Median predicted punishment is 3 years, while more than a third of the punishments are below 15 months.}

\label{fig:hist_punishment}
\end{figure}

\paragraph{Identifying sentencing patterns in sexual assault cases in Israel.} 
Using our best performing model, we present rough statistics regarding the punishments given in the past thirty years in Figure ~\ref{fig:hist_punishment}. The median sentenced punishment throughout all legal decisions of our data is 36 months, however we observe most commonly punishments are under a year. While the sentencing decisions are generally available in legal search engines, annotating them is an expensive process. For this reason many statistical observations are hard to obtain. This demonstrates the potential contribution of our task from a socio-legal point of view.

\section{Related Work}

Most related to our work are those that bring together domain expertise with ML models to extract information from specialized texts. This is the case for \newcite{soh-etal-2019-legal} who showed that conditional random fields perform better than DNN for sentence border detection for the legal domain. While this is considered a closed problem in NLP they showed that this is not the case for legal texts. Similar paradigms can be found in the medical field, where domain knowledge also plays a crucial role.  \newcite{malmasi2019comparison} show that in some cases a rule-based approach achieves better performances than SVM. 
\newcite{2020legalBERT} recently introduced a language model fine-tuned superficially for the English legal domain. Taking a similar approach for the APE task is an interesting avenue for future work.

\section{Conclusions}
In this work we created the first annotated corpus of Hebrew language sentencing decisions, focusing on sexual assaults.
We compared a rule-based approach with supervised learners using the unique attributes of the legal language for representing sentences. We found that the rule-based approach achieved best results with an average error rate of 5 months and accuracy of 68\% in extracting the punishment sentence. 
Our analysis shows that such research could focus on fine tuning of the supervised models. While supervised learning models help us narrow down a full legal document to 2 - 5 sentences that include the punishment, further research can contribute in reaching a single target sentence, which could also benefit from our error analysis, especially regarding the probation sentences, perhaps targeting them separately in a prior task and using them as features. 

\section*{Acknowledgments}
We thank the anonymous
reviewers for their helpful comments and feedback.
This work was supported in part by a research gift
from the Allen Institute for AI and by a research grant from the Center for Interdisciplinary Data Science Research (CIDR) at the Hebrew University of Jerusalem. 

\bibliography{anthology,custom}
\bibliographystyle{acl_natbib}

\end{document}